\documentclass[letterpaper]{article} % DO NOT CHANGE THIS
\usepackage{aaai25}  % DO NOT CHANGE THIS
\usepackage{times}  % DO NOT CHANGE THIS
\usepackage{helvet}  % DO NOT CHANGE THIS
\usepackage{courier}  % DO NOT CHANGE THIS
\usepackage[hyphens]{url}  % DO NOT CHANGE THIS
\usepackage{graphicx} % DO NOT CHANGE THIS
\usepackage{natbib}  % DO NOT CHANGE THIS AND DO NOT ADD ANY OPTIONS TO IT
\usepackage{caption} % DO NOT CHANGE THIS AND DO NOT ADD ANY OPTIONS TO IT
\usepackage{algorithm}
\usepackage{algorithmic}
\usepackage{xcolor}

\usepackage{newfloat}
\usepackage{listings}

\usepackage{booktabs}
\usepackage{amsmath}
\usepackage{amssymb, amsfonts}

\DeclareCaptionStyle{ruled}{labelfont=normalfont,labelsep=colon,strut=off} % DO NOT CHANGE THIS
\floatstyle{ruled}
\newfloat{listing}{tb}{lst}{}
\floatname{listing}{Listing}
\title{Unsupervised Translation of Emergent Communication}

\author{
    Ido Levy\textsuperscript{\rm 1}, 
    Orr Paradise\textsuperscript{\rm 2}, 
    Boaz Carmeli\textsuperscript{\rm 1}, 
    Ron Meir\textsuperscript{\rm 1}, 
    Shafi Goldwasser\textsuperscript{\rm 2}, 
    Yonatan Belinkov\textsuperscript{\rm 1}
}
\affiliations {
    \textsuperscript{\rm 1}Technion – Israel Institute of Technology, Haifa, Israel\\
    \textsuperscript{\rm 2}University of California, Berkeley, CA, USA\\
    \{idolevy, boaz.carmeli\}@campus.technion.ac.il, rmeir@ee.technion.ac.il, belinkov@technion.ac.il,\\
    \{orr.paradise, shafi.goldwasser\}@berkeley.edu
}

\begin{document}

\maketitle

\begin{abstract}

Emergent Communication (EC) provides a unique window into the  language systems that emerge autonomously when agents are trained to jointly achieve shared goals. However, it is difficult to interpret EC and evaluate its relationship with natural languages (NL).
This study employs unsupervised neural machine translation (UNMT) techniques to decipher ECs formed during referential games with varying task complexities, influenced by the semantic diversity of the environment. 
Our findings demonstrate UNMT's potential to translate EC, illustrating that task complexity characterized by semantic diversity enhances EC translatability, while higher task complexity with constrained semantic variability exhibits pragmatic EC, which, although challenging to interpret, remains suitable for translation. This research marks the first attempt, to our knowledge, to translate EC without the aid of parallel data.

\end{abstract}

\begin{figure*}[ht]
    \centering
    \includegraphics[width=0.8\linewidth]{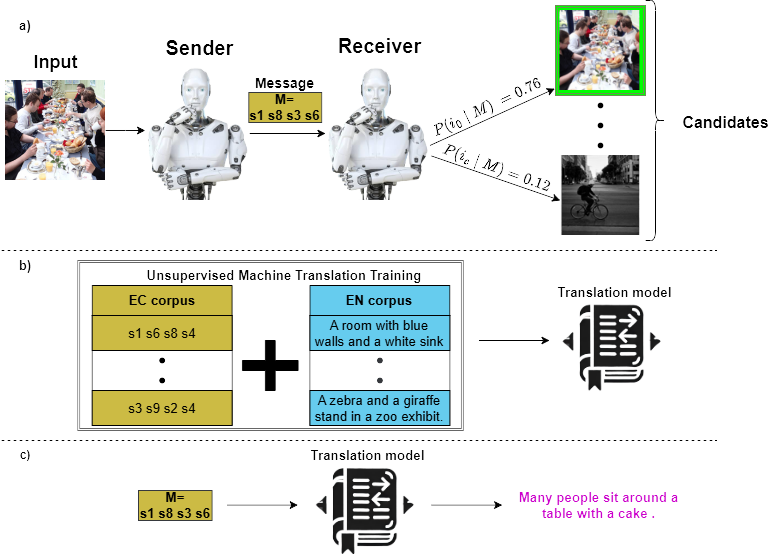}

    \caption{\textbf{(a)} Illustration of the referential game setup. The Sender observes an image and sends a message to the Receiver, who must identify the correct image from a set of candidates based on the message received. The exchanged messages are recorded to create the EC corpus. \textbf{(b)} Using the monolingual EC corpus and a monolingual English caption corpus to train the UNMT system. \textbf{(c)} The UNMT translating an EC message into English. 
    }
    \label{fig:referential_game}
\end{figure*}

\section{Introduction}
Emergent communication (EC) describes the phenomenon in which AI agents develop communication protocols to achieve shared goals \cite{giles2003learning, 5045926, boldt2024review, brandizzi2023toward}. This capacity has garnered attention due to its significant potential for understanding the complexities of language formation and evolution within multi-agent systems. Yet, ECs remain largely opaque,  difficult to interpret and translate into human-readable forms.

Although efforts to understand EC have been made,  interpretability remains elusive. Traditional approaches, such as \textit{topographic similarity} \cite{brighton2006understanding}, measure the correlation between message and input distances, provide a coarse measure of EC structures, but do not capture the full nuances of EC compositionality. Recent advancements aim to use natural language (NL) to analyze EC \cite{NEURIPS2022_9f9ecbf4, carmeli2024concept, Chaabouni2020CompositionalityAG}. 
Another line of research aims to translate EC into NL using parallel data (EC-NL pairs) to gain a more intuitive understanding \cite{andreas_etal_2017_translating,yao2022linking}. However, this approach may introduce certain biases in the translation process.

Our approach is to utilize \textit{unsupervised neural machine translation} (UNMT) to translate the emergent languages developed during referential games of varying complexity. UNMT is particularly suited for this task as it does not require parallel data \cite{Lample2018, artetxe2018robust, conneau2017word, lample2018phrase, xu2018unsupervised}, which is typically unavailable for AI-generated languages. To facilitate translation, we generate an EC corpus from each game by compiling the messages exchanged between agents. Concurrently, we employ a separate English caption dataset as a linguistic prior, which provides a resource of image descriptions for images provided to the agents during the game (Naturally,  the images provided to the agents contains no caption). By integrating these distinct datasets, we can adapt existing UNMT techniques \cite{chronopoulou_etal_2020_reusing} to translate the emergent communications into English, showcasing their translatability across varying task complexities.

\begin{figure*}[ht]
    \centering
    \includegraphics[width=0.9\linewidth]{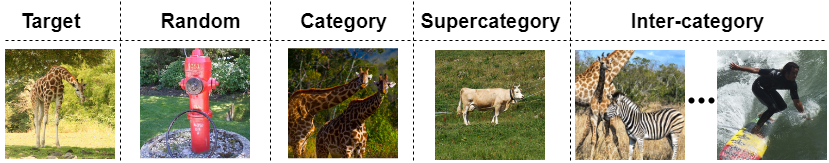}
    \caption[Game Complexity in Referential Games]{Illustration of various levels of game complexity in referential games. The target image, a giraffe, is shown alongside different levels of distractors: a red fire hydrant representing an random game; another giraffe for category discrimination;
    % \orrp{The preceding sentence is confusing to me.};
    cow in a lush field for supercategory discrimination; and a zebra alongside a giraffe and a person wavesurfing for a Inter-category game, where overlapping concepts (giraffes) illustrate the images' inherent complexity in a multi-category setting. Each column exemplifies the escalation of game complexity and the corresponding increase in potential target and distractor candidates.}
    \label{fig:discrimination_games_complexities}
\end{figure*}

Leveraging the capabilities of UNMT, our study dives into the intricacies of structured referential games of varying complexity \cite{lewis2008convention, lazaridou2016multi, choi2018compositional, guo2019emergence}, including \textit{Random}, \textit{Inter-category}, \textit{Supercategory}, and \textit{Category} image discrimination tasks. 
In these games, agents must identify a target image, such as a \textit{giraffe}, from a set of distractors. Distractors are non-target images selected based on the game type, adding complexity to the task (Figure \ref{fig:referential_game}a).
% \shafi{did the term "distractors" ever get defined?}
For example, in a Supercategory game, a distractor for a \textit{giraffe target} can be \textit{cow}, whereas in a Random game, the distractors could be as unrelated as a \textit{fire hydrant} (Figure \ref{fig:discrimination_games_complexities}), for the same \textit{giraffe} target. These games are deliberately designed to simulate diverse communication environments, aimed at constructing ECs that increasingly resemble human languages, and will eventually be useful for interpretability and usability of AI-generated languages in real-world scenarios. We hypothesize that, akin to the evolution of human languages that adapt in response to social and environmental pressures \cite{kuhl2004early, kottur2017natural}, the complexity inherent in these games will catalyze similar transformations in the EC.

Through our research, we have established a new benchmark for translating EC into NL. This benchmark leverages a comprehensive suite of metrics designed to capture the intricacy and diversity of AI-generated languages. These metrics assess the translation quality in several dimensions: standard machine translation metrics with image captions as ground truth, semantic correlation with images using CLIP scores \cite{hessel_etal_2021_clipscore} to assess interpretability, and intrinsic text analysis through metrics like token-type ratio that reflect linguistic richness in the translated text.

Our experiments support the potential of UNMT to translate AI-generated languages into human-readable text. Notably, the \textit{Inter-category} setting showcased superior translation quality, evidenced by higher BLEU and METEOR scores. Furthermore, our analysis indicates that greater vocabulary usage, reflecting a broader range of vocabulary in EC messages, and increased entropy, signaling message unpredictability, may pose challenges to translation accuracy.
Qualitative analysis of the resulting translations suggests that UNMT successfully captures the main objects or themes in the described images, but not all the fine-grained details. 
A notable discovery of our research is the lack of correlation of translation performance across ECs originating from different setups, whether due to varying game complexities or initialization seeds for randomness, when tested on identical test sets.
This lack of correlation indicates that each instantiation of a game cultivates distinct communication protocols and not a single universal standard.

Our work makes three key contributions:
\begin{itemize}
  \item We provide a novel application of UNMT techniques to decipher AI agents' EC without the need for parallel data.
  \item We investigate how the complexity of referential games influences the ECs development and their translatability.
  \item We establish a novel benchmark for evaluating EC translations by introducing a comprehensive set of metrics tailored to assess the sophistication and variability of ECs.
\end{itemize}

\section{Related Work}
Research in EC has developed various methodologies to understand and evaluate the structure and sophistication of languages that emerge among AI agents. Early efforts to quantify communication among agents focused on traditional metrics like \textit{topographic similarity}, which assesses the correlation between the distances of messages and their corresponding inputs \cite{brighton2006understanding}. To understand EC, \citet{Lazaridou2018EmergenceOL} explored the input space effect on EC by demonstrating that AI agents can develop EC that solves the shared task using both symbolic and pixel inputs, while \citet{lee2018emergent} highlighted how factors like model capacity and communicative bandwidth foster systematic compositional structures in EC.
To understand NL properties in EC, several studies have explored aspects such as Zipf's Law of Abbreviation \cite{ueda2021relationship} and Harris's Articulation Scheme \cite{ueda2022word}. Additional work focused on compositionality, examining disentanglement-based measures \cite{Chaabouni2020CompositionalityAG}, Adjusted Mutual Information (AMI) \cite{mu2021emergent}, and rigorous methods to verify compositional structures \cite{Vani2021IteratedLF, andreas2019measuring}.
More recent advances have introduced more nuanced metrics that aim to understand EC using NL. For instance, \citet{NEURIPS2022_9f9ecbf4} evaluated learned EC on unseen test data to assess generalization, providing insights into NL aspects. \citet{carmeli2024concept} proposed mapping EC symbols to NL concepts, assessing EC's compositionality. However, their approach forms a global mapping of atomic symbols, rather than full translation of individual messages.  Most closely related to our approach is work translating EC into human-understandable language.  \citet{andreas_etal_2017_translating}  translated continuous communication channels into NL by collecting agents' messages and corresponding NL strings within the same games. \citet{yao2022linking} trained image discrimination referential game and translated the EC to NL captions grounded on the same images. While promising, these studies crucially used parallel EC-NL pairs for training the translation system. In contrast, our approach does not rely on any parallel data.
Several works leverage EC as a tool to enhance NL translation in unsupervised settings \cite{downey_etal_2023_learning, lee2018emergent}. However, unlike our focus on directly translating EC, these studies utilized EC to facilitate future NL translation tasks.

\citet{guo2021expressivity} termed EC expressivity as the amount of input information encoded in EC, defining theoretically and demonstrating empirically the impact of unpredictability and contextual similarity on EC expressivity. In this paper, we extend the concept of contextual complexity by introducing  a broader spectrum of complexity, including low, moderate, and high complexity levels. This allows us to systematically explore how varying degrees of similarity among distractors influence the translatability of EC.

\section{Complexity of Referential Games}

Complexity refers to the degree of similarity between distractors and the target in referential games. This property can range from being semantically related to completely unrelated. In this section, we define four levels of complexities that represent a wide range of game complexity levels.

\subsection{Preliminaries} 
We focus on a two-agent setup, consisting of a \textit{Sender} and a \textit{Receiver}, operating under predefined rules within simulated environments. These agents interact through a discrete communication channel \cite{lazaridou2020emergent, denamganai2020referentialgym} to successfully complete shared tasks by conveying information about objects or concepts represented in the environment.

In the framework of referential games, the \textit{Sender} observes an image \( i \in \mathcal{D} \) and sends a message \( m \in \mathcal{M} \) to describe the image to the \textit{Receiver}. The \textit{Receiver} uses this message to identify the correct image from a set of candidates, where the target's position is randomized. This setup is defined by the tuple \( (A, \mathcal{M}, \mathcal{D}, \mathcal{C}, \mathcal{S}) \), where:
\begin{itemize}
  \item \( A = \{A_S, A_R\} \) are the Sender and Receiver agents.
  \item \( \mathcal{M} \) represents the message space.
    \item \( \mathcal{D} \) is the dataset used to sample images.
    \item \( \mathcal{C} \) is the set of categories in the dataset.
    \item \( \mathcal{S} \) encompasses the set of supercategories, grouping similar categories together.
    
\end{itemize}

\subsection{Game Complexity Levels} \label{ssec:complexities_definitions}

We craft the environment for EC games around a target image \(i_{T}\) classified under a category \(c_T\) and a supercategory \(s_T\). The distractor environment is composed of \(d\) images chosen based on the following criteria:

\begin{enumerate}
\item \textbf{Random:}
In this simplest form of the game, agents operate in a completely random environment, where each distractor image $i$ is sampled uniformly from the dataset:  $i \sim \mathcal{D} $. For example, the task could involve distinguishing between a car and a zebra. This basic level challenges the agents to develop fundamental communication protocols from scratch, testing their ability to generate and understand rudimentary language. In \citet{Lazaridou2018EmergenceOL}, it is described as a \textit{uniform} game.

\item \textbf{Category Discrimination:}
The most contextually complex level involves agents discriminating between images containing objects of the same category, such as distinguishing between different images of giraffes. Here each distractor $i$ is sampled from the set of images that share a category with that of the target images:  
$ i \sim \mathcal{D}_{c_T}$, where 
$\mathcal{D}_{c_T} = \{ i \in \mathcal{D} \mid c_T \in \text{categories}(i) \}$.  
This level of complexity is designed to encourage agents to develop highly sophisticated and detailed communication strategies, which means the agents must use nuanced and precise EC to describe subtle differences. In \citet{guo2021expressivity}, it is described as a \textit{high-complexity source}.

\item \textbf{Supercategory Discrimination:}
At this level, agents must distinguish between semantically close images within the same broad category. 
It is defined analogously to the previous game: 
$ i \sim \mathcal{D}_{s_T}$, where 
$\mathcal{D}_{s_T} = \{ i \in \mathcal{D} \mid s_T \in \text{supercategories}(i) \}$.
For instance, they might need to differentiate between a giraffe and a cow in the `animal' supercategory. This scenario emphasizes the importance of context-aware communication, encouraging agents to address and articulate subtle differences in features and attributes.
It is similar to the \textit{context-dependent} game described by 
 \citet{Lazaridou2018EmergenceOL}.

\item \textbf{Inter-category:} 
In this game setup, $d$ categories are sampled \emph{without replacement}:
$ \{ c_1, \ldots, c_d \} \subseteq \mathcal{C} \setminus c_T$. Then $d$ distractors are sampled, one from each category:  $\forall j \in \{1, \dots, d\}, \quad i_j \sim \mathcal{D}_{c_j}$. 
Agents discriminate between images that, while representing different categories, may contain overlapping features due to the dataset's inherent complexity. An example task might involve distinguishing between bike and tennis scenes, both featuring a person. This level introduces a layer of ambiguity, requiring agents to refine their communication to highlight distinguishing features in overlapping contexts.

\end{enumerate}

\section{Methodology}

The training regime of our methodology consists of two phases. First, we train the agents on image discrimination games (Figure \ref{fig:referential_game}a) of varying complexity (Section \ref{ssec:complexities_definitions}) and record the exchanged messages. Each game serves as a unique setup to provoke distinct communication strategies among agents, depending on task complexity. Our next phase is to use EC messages from each game as a monolingual EC corpus for training UNMT system (Figure \ref{fig:referential_game}b).

\paragraph{UNMT Architecture}
For details on UNMT techniques foundational to our study,  see Appendix \ref{ssc:Unsupervised_Machine_Translation}.
Our work employed the UNMT system by
\citet{chronopoulou_etal_2020_reusing}
that is implemented in three steps:
\begin{enumerate}
    \item \textbf{Pre-training:} Utilizing a high-resource English corpus to train our model, ensuring a rich linguistic foundation.
    \item \textbf{Fine-tuning:} Adapting the pre-trained model using EC data collected from AI agents, enhancing its ability to handle the specific linguistic features of EC. This phase creates a shared embedding space with the EC.
    \item \textbf{Back-translation and Denoising:} Employing back-translation and denoising techniques to refine the model's output and improve translation between EC and English.
\end{enumerate}

\section{Experimental Setup}

\subsection{Data Specifications}

For this study, we employed the MSCOCO dataset \cite{Lin2014MicrosoftCC}, a diverse collection of 117K complex images that are annotated with various NL concepts. Each image in the dataset is paired with five distinct captions, providing high-quality captions to inject prior knowledge of image descriptions while training our UNMT models, and to serve as valuable reference points for evaluating translation performance. More information about the dataset appears in Appendix \ref{sec:appendix_dataset}.

\subsection{AI Agents Architecture}

AI agents were configured with a hybrid architecture combining elements of LSTMs \cite{hochreiter1997long} and  ResNets \cite{koonce2021resnet}, to process images and to generate ECs effectively, representing a typical architectural standard in the EC domain. Each agent was initialized with different seeds to ensure the robustness and generalizability of the results. The Sender and the Receiver are sharing the ResNet weights, to ensure that both agents process visual information consistently, supporting coherent communication across the system by minimizing discrepancies in visual perception. The ResNet was initialized with pre-trained weights from ImageNet \cite{ImageNet}. The joint training objective is infoNCE \cite{Oord2018RepresentationLW} which is often used in discrimination setups.

\setlength{\tabcolsep}{1mm} % Adjust column spacing
\renewcommand{\arraystretch}{0.9} % Adjust row spacing
\begin{table*}[t]
\centering
% \small % Optional: Reduces font size for the table
\begin{tabular}{lccccc}
\toprule
\textbf{Metric} & \textbf{Category} & \textbf{Supercategory} & \textbf{Random} & \textbf{Inter-category} & \textbf{Baseline}\\
\midrule
ACC 2 (\%) & 93.62 {\scriptsize\(\pm\) 0.20} & 96.35 {\scriptsize\(\pm\) 0.56} & 96.89 {\scriptsize\(\pm\) 0.50} & 96.90 {\scriptsize\(\pm\) 0.46} & 54.25 {\scriptsize\(\pm\) 0.81} \\
ACC 10 (\%) & 70.51 {\scriptsize\(\pm\) 0.88} & 77.35 {\scriptsize\(\pm\) 2.96} & 80.20 {\scriptsize\(\pm\) 2.25} & 78.40 {\scriptsize\(\pm\) 2.62} & 13.55 {\scriptsize\(\pm\) 1.23} \\
VU (\%) & 55.95 {\scriptsize\(\pm\) 2.04} & 73.10 {\scriptsize\(\pm\) 9.77} & 68.26 {\scriptsize\(\pm\) 9.03} & 66.34 {\scriptsize\(\pm\) 7.73} & 100.0 {\scriptsize\(\pm\) 0.0} \\
Entropy & 6.78 {\scriptsize\(\pm\) 0.41} & 9.27 {\scriptsize\(\pm\) 1.38} & 7.79 {\scriptsize\(\pm\) 0.80} & 7.71 {\scriptsize\(\pm\) 2.00} & 14.34 {\scriptsize\(\pm\) 1.02} \\
Novelty (\%) & 0.84 {\scriptsize\(\pm\) 0.42} & 6.85 {\scriptsize\(\pm\) 2.28} & 1.35 {\scriptsize\(\pm\) 0.66} & 3.20 {\scriptsize\(\pm\) 1.46} & 0.0 {\scriptsize\(\pm\) 0.0} \\
\bottomrule
\end{tabular}
\caption{EC results, segmented by complexity and including a baseline of random agent communication. Metrics reported as mean \(\pm\) standard error from 5 seeds. ACC 2 and ACC 10 represent discrimination accuracy with 1 and 9 distractors, respectively.}
\label{tab:ec_game_results}
\end{table*}

\subsection{Implementation Details} \label{Implementation_details}
We used the EGG framework \cite{kharitonov_etal_2021}
to train the models with a batch size of 1024 and an initial learning rate of 0.001, using the Adam optimizer. Training epochs were set to 50, with early stopping based on validation loss. In each game, nine distractors are sampled based on the complexity policy. The same target images were used for each complexity's test, but a new set of distractors was sampled according to the complexity of the game.
The agents' communication channel is quantized \cite{carmeli2023emergent},  consisting of 64 symbols, represented as binary vectors of length 6, with each message composed of 6 symbols and \textit{EOS} symbol. To ensure robustness, each referential game was run with 5 different random seeds.

The UNMT model utilized a pre-trained XLM, which was fine-tuned on both the EC corpus and MSCOCO captions. 
More details on the EN corpus are provided in Appendix \ref{ssc:appendix_captions_statistics}.
See Appendix  \ref{sec:appendix_hyperparameters}
for full hyperparameters.

\setlength{\tabcolsep}{1mm} % Adjust column spacing
\renewcommand{\arraystretch}{0.9} % Adjust row spacing
\begin{table*}[ht]
    \centering
    % \small % Optional: Reduces font size for the table
    \begin{tabular}{lcccccc}
        \toprule
        \textbf{Metric} & \textbf{Category} & \textbf{Supercategory} & \textbf{Random} & \textbf{Inter-category} & \textbf{Baseline} & \textbf{Chinese} \\
        \midrule
        Novelty (\%) & 58.74 {\scriptsize\(\pm\) 7.81} & \textbf{70.00} {\scriptsize\(\pm\) 1.68} & 60.54 {\scriptsize\(\pm\) 4.25} & 57.36 {\scriptsize\(\pm\) 5.83} & 100.0 & 88.19 \\
        BLEU Score & 7.41 {\scriptsize\(\pm\) 0.47} & 6.08 {\scriptsize\(\pm\) 0.31} & 6.85  {\scriptsize\(\pm\) 0.34} & \textbf{9.21} {\scriptsize\(\pm\) 0.45} & 0.071 & 5.65 \\
        BERTScore & \textbf{0.734} {\scriptsize\(\pm\) 0.001} & 0.730 {\scriptsize\(\pm\) 0.001} & 0.729  {\scriptsize\(\pm\) 0.001} & 0.730 {\scriptsize\(\pm\) 0.001} & 0.543 & 0.74 \\
        METEOR Score & 0.295 {\scriptsize\(\pm\) 0.06} & 0.276 {\scriptsize\(\pm\) 0.06} & 0.289 {\scriptsize\(\pm\) 0.06} & \textbf{0.310} {\scriptsize\(\pm\) 0.07} & 0.115 & 0.234 \\
        ROUGE-L & 0.361 {\scriptsize\(\pm\) 0.001} & 0.343 {\scriptsize\(\pm\) 0.006} & 0.352 {\scriptsize\(\pm\) 0.003} & \textbf{0.370} {\scriptsize\(\pm\) 0.002} & 0.173 & 0.356 \\
        Jaro Similarity & 0.678 {\scriptsize\(\pm\) 0.02} & 0.673 {\scriptsize\(\pm\) 0.02} & 0.676 {\scriptsize\(\pm\) 0.02} & \textbf{0.682} {\scriptsize\(\pm\) 0.02} & 0.601 & 0.674 \\
        CLIP Score & 0.180 {\scriptsize\(\pm\) 0.018} & 0.176 {\scriptsize\(\pm\) 0.019} & 0.183  {\scriptsize\(\pm\) 0.020} & \textbf{0.191} {\scriptsize\(\pm\) 0.019} & 0.151 & - \\
        TTR (\%) & 0.42 {\scriptsize\(\pm\) 0.05} & \textbf{0.71} {\scriptsize\(\pm\) 0.14} & 0.58 {\scriptsize\(\pm\) 0.11} & 0.59 {\scriptsize\(\pm\) 0.15} & 0.19 & 0.83 \\    
        \bottomrule
    \end{tabular}
    \caption{UNMT Performance Across Different Game Complexities. Each metric is reported as mean \(\pm\) standard error, derived from 3 different ECs that emerged from the same complexity.}
    \label{tab:unmt_overall_performance}
\end{table*}

\subsection{Evaluation Metrics}

To comprehensively assess the effectiveness of EC games and the UNMT, we employed a diverse set of evaluation metrics.
See Appendix \ref{ssec:appendix_evaluation_metrics_EC} 
for more details on the metrics.

\subsubsection{EC Games Metrics}
We used the following metrics to evaluate the performance and sophistication of the EC emerged by AI agents during the referential games:
\begin{itemize}
    \item \textbf{Game Accuracy}: Measures the percentage of correct identifications made by the agents \cite{lewis2008convention}.
    \item \textbf{Vocabulary Usage (VU)}: Assesses vocabulary diversity by measuring the range of unique symbols utilized during game sessions \cite{graesser2003automated}.
    \item \textbf{Message Entropy}: Calculates the uncertainty in the agents' messages \cite{shannon1948mathematical}.
    \item \textbf{Message Novelty}: Determines whether test messages differ from training messages \cite{chomsky2002syntactic}.
    \item \textbf{Topographic Similarity (TopSim)}: Measures the correlation between pairwise distances in the input space and the corresponding distances in the message space, using Euclidean distance for input embeddings and edit distance for messages. \cite{brighton2006understanding}
    \item \textbf{Disentanglement Metrics.} Bag-of-Symbols (BosDis) and Positional (PosDis) \cite{chaabouni_etal_2020_compositionality} both measure how distinctly attributes are encoded. BosDis computes per-symbol mutual information (MI) gaps between primary and secondary attributes, normalized by symbol entropy, while PosDis applies the same approach across positions in a message.
    \item \textbf{Adjusted Mutual Information (AMI)}: Assesses the alignment between messages and underlying concepts \cite{mu2021emergent}. In the multi-label case, \textbf{mAMI} extends AMI by averaging per-concept scores.

\end{itemize}

\subsubsection{UNMT Metrics} \label{ssec:UNMT_metrics}
In the absence of parallel data, we used the captions of the images to evaluate the translation quality of UNMT. Since each caption captures different nuances and writing styles, and we aim to translate EC messages generated by the agents to a specific image, a good translation would be similar to at least one of the captions. Therefore, we reported the max score over 5 captions to account for this variability. Metrics were calculated using the  \texttt{string2string} package \cite{Suzgun2023string2stringAM}.  We employed a variety of metrics that capture both exact match accuracy and semantic alignment:
\begin{itemize}
    \item \textbf{BLEU}: Measures the n-gram overlap between translated text and reference captions \cite{papineni2002bleu}.  We employed the default SacreBLEU \cite{SacreBLUE2018}, which provides a standardized way to evaluate translations quality by ensuring consistent and reproducible scores. 
    \item \textbf{METEOR}: Considers precision, recall, synonymy, stemming, and word order \cite{banerjee2005meteor}.
    \item \textbf{ROUGE-L}: Focuses the longest common subsequence between the translated text and reference \cite{lin2004rouge}.
    \item \textbf{BERTScore}: Utilizes contextual embeddings from BERT to evaluate semantic similarity between the translated EC and reference captions \cite{zhang2019bertscore}.

    \item \textbf{CLIP Score}: Employs the CLIP model to assess the semantic alignment between the translated text and the corresponding image. This metric introduces a novel approach for evaluating how well the translation reflects the content and context of the image that prompted the original EC message \cite{hessel_etal_2021_clipscore}.

    \item \textbf{Jaro Similarity}: Measures similarity and character transpositions
    between texts \cite{jaro1989advances}.
    \item \textbf{Text-Type Ratio (TTR)}: Analyzes the lexical diversity within the message translations \cite{richards1987typeratio}.
    \item \textbf{Novelty Score}: Evaluates how many unique n-grams are produced in translations compared to the training corpus.

\end{itemize}

\section{Results}

\subsubsection{Emergent Communication Games}

The baseline performs marginally above random,  which is suggestive of \citet{buzaglo2024uniform}'s findings on limited generalization in randomly sampled networks. As shown in Table \ref{tab:ec_game_results}, it significantly underperforms across all metrics, with a VU of 100\%, indicating an arbitrary use of the EC vocabulary without solving the task. These results demonstrate the effectiveness of the EC strategies developed in the games.

As expected, in the ten-candidates (ACC 10) scenario, the game accuracy scores degrade with increasing task difficulty, with \textit{Category} game complexity presenting significantly lower results with 70.5\%. Despite differences in the communication channel, as opposed to \citeauthor{guo2021expressivity}
, who claim that more contextual similarity leads to higher expressivity, we observe that the Category game complexity, which emphasizes fine-grained distinctions, has the lowest VU (55.95\%), a metric indicating EC richness. We attribute this finding to the complexity level, as more specificity is required, agents might use what we hypothesize as ``low-level features'', like angle, background, and object size, which require fewer symbols for the task, leading to the development of simplistic communication strategies.

The low standard errors across the accuracy metrics indicate that EC assessment metrics such as \textit{VU} and entropy do not correlate well with classic referential game accuracy metrics. This indicates that the emerged protocols successfully solve tasks using diverse strategies. The overall low novelty scores across games, combined with low standard errors, imply that agents largely rely on memorized messages generated during training. The table shows some ambiguity in the results due to interval overlap in some metrics across different complexities. However, there is a high correlation between Entropy and \textit{VU} ($\rho = 0.85$) across all complexities.\footnote{Reported correlations are significant at $p < 0.05$.} Among the \textit{Supercategory}, \textit{Random}, and \textit{Inter-category} complexities, we observe that \textit{Supercategory} possesses the highest scores in both Entropy and \textit{VU}, indicating a need for a more extensive vocabulary to discriminate between semantically related concepts. Despite the overall low novelty scores, \textit{Supercategory}'s is significantly higher at 6.85\%, which is directly correlated with Entropy ($\rho = 0.8$) and \textit{VU} ($\rho = 0.7$). These correlations reveal that rich and diverse EC impacts the novelty of the agents' messages.

\begin{figure}[t] 
    \centering
    \includegraphics[width=1\linewidth]
        {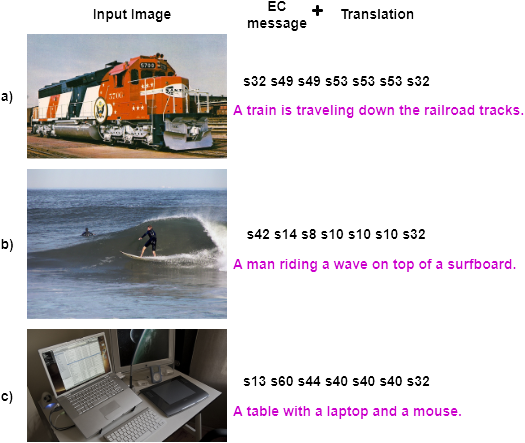}
    \caption{
        Selected translation examples. Each panel shows an EC message, composed of six symbols followed by an \textit{EOS} symbol, paired with its corresponding translation beneath. The translations capture nuanced details from the visuals, producing coherent and contextually appropriate text. Notably, in panel c), the MT model extends its response beyond the visible elements in the image, by adding ``\textit{and a mouse}'', suggesting a tendency to ``hallucinate'' details potentially influenced by prior context knowledge and previous training on similar caption lengths.
        }
    \label{fig:KS_translation_examples}
\end{figure}

\subsubsection{Unsupervised Machine Translation}

\paragraph{Benchmark UNMT system's performance} We conducted two experiments to understand the UNMT capabilities. The first experiment aimed to calibrate the performance of our UNMT approach on a real, high-resource language by translating Chinese captions to English. For this experiment, we used MSCOCO-CN \cite{li2019coco} composed of $\sim 30$K annotated images. The results, detailed in Appendix \ref{sec:appendix_chinese_translation}
, showed fluent English translations with an overall BLEU score of 5.65 points. The second experiment, serves as a baseline, evaluates the performance of the UNMT on EC generated by random agents that did not undergo the training phase designed to optimize EC. The results, presented in Table \ref{tab:unmt_overall_performance}, indicate significantly lower performance metrics, with BLEU scores near zero. This confirmed that random agent communication did not form meaningful linguistic patterns, strengthening our translation reliability on EC that emerged from AI agents' collaboration towards a shared goal.

\paragraph{UNMT of EC} Our findings demonstrate the potential of UNMT for translating EC into English without the use of parallel data. This  is reflected in all translation metrics: While modest compared to full-fledged supervised translation of natural languages, these metrics  are significantly higher than the baseline, indicating the formation of meaningful EC during collaborative training. For illustrative translation examples,  refer to Figure \ref{fig:KS_translation_examples} and Appendix \ref{sec:appendix_king_solomon_translations}.
Interestingly, these scores are similar or slightly better than the Chinese-to-English translation results, which could be attributed to the richness differences between EC and Chinese or to the relatively small number of Chinese captions compared to EC messages. The translations show a considerable level of coherence and meaning. Notably, the \textit{Inter-category} complexity achieved the highest BLEU score (9.21) and ROUGE-L score (0.37). We speculate this is due to the fact that in this setup the image discrimination game involves different categories, shaping the information conveyed by the EC to be specific to a category among others that may share mutual features, making it more concept-related, and therefore aligning the EC more towards NL, facilitating translations compared to other complexity levels. This alignment is supported by empirical evidence in Appendix \ref{sec:appendix_ec_classification_2_NL_concepts},
demonstrating successful mapping of EC messages to NL concepts. Evaluating translation-image alignment, the CLIP score reaches 0.19, compared to 0.29 for MSCOCO's actual captions and 0.15 for the random baseline, indicating acceptable performance given CLIP’s tendency toward lower scores with diverse, complex images. Thus, while our translations are better than translating random messages, they fall short of aligning real captions. 

While the translations are coherent, we observed an interesting phenomenon where translations sometimes capture several features from the image but end with unrelated ones. We attribute this to the model ``hallucinating'' based on the narrow distribution of caption lengths it encountered during training; for an example refer to Figure \ref{fig:KS_translation_examples}c. This suggests the model captures multiple concepts from the messages, but occasionally extends the translation with irrelevant details. This observation indicates that while the translations demonstrate potential and capture the main theme, they are not perfect and warrant further investigation in future research.

\begin{figure}[ht]
    \centering
    \includegraphics[width=1\linewidth]{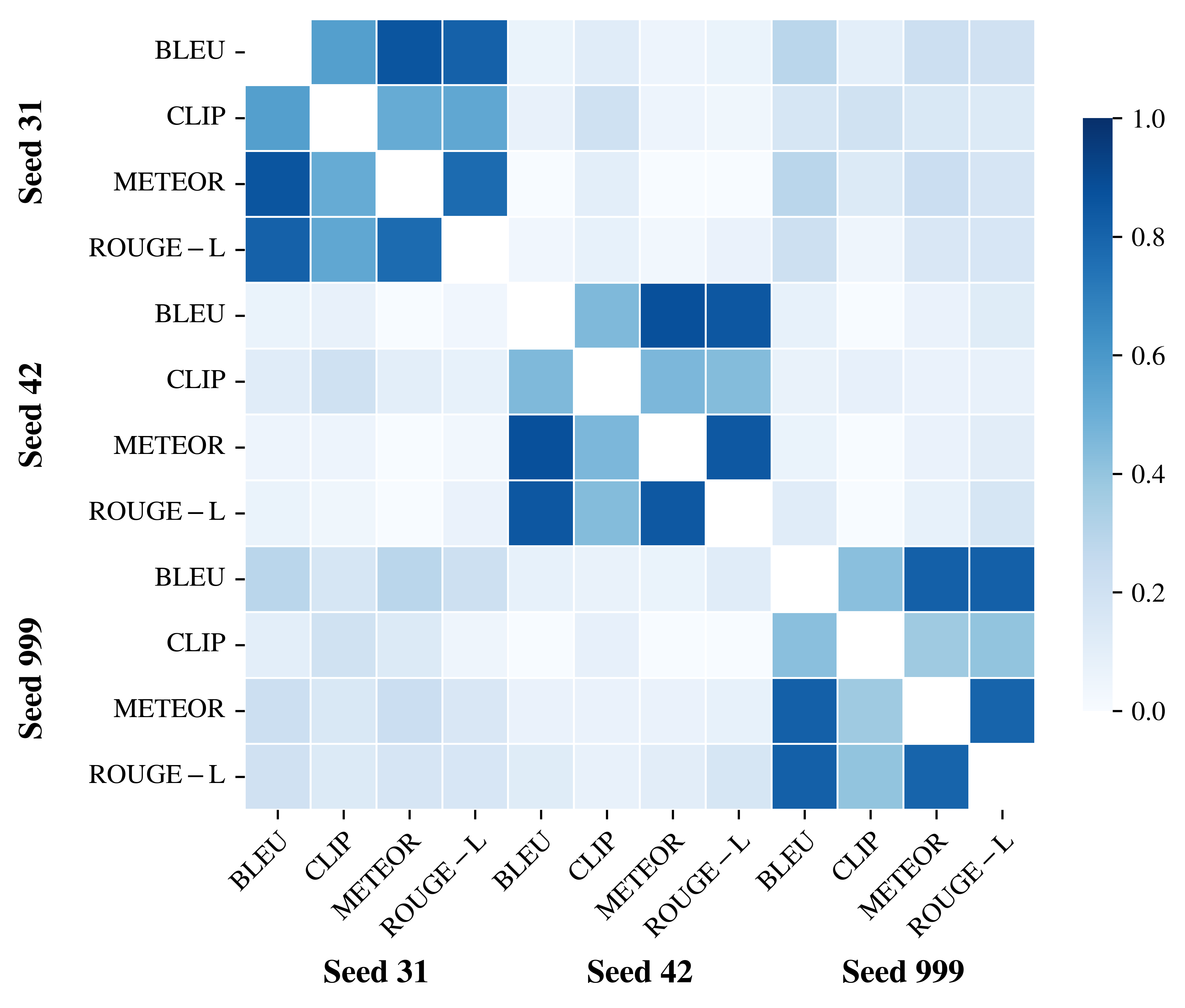}
    \caption{Correlation matrix for Inter-Category complexity across different seeds. The near-zero correlation between seeds indicates that translation model performs uniquely on the same examples. Additionally, the strong correlation between exact match metrics, which positively correlate with the semantic metric within each seed, highlights the significant text-image alignment achieved by our translations.}
    \label{fig:correlation_matrix_one_column}
\end{figure}
\paragraph{Game complexity vs. Translatability}
Several key insights emerge from our analysis. Translation performance metrics, including BLEU and ROUGE-L, are highest for \textit{Inter-category}. This indicates that partial supervision and candidates from broad categories facilitate better translation, making ECs in these contexts more translatable.
In contrast, \textit{Supercategory}, with high contextual complexity, presents the highest novelty score (70.00\%) while having the lowest BLEU (6.08) and ROUGE-L (0.352). This highlights the difficulty in translating ECs that were developed based on semantically similar image discrimination. As illustrated by Table \ref{tab:ec_game_results}, this complexity type is associated with the highest Entropy and VU, indicating that agents communicated diverse messages, which directly affected translation performance and encouraged more novel translations.
Conversely, the \textit{Category} complexity, designed for fine-grained distinctions, achieved moderate translation metrics with BLEU (7.41) and ROUGE-L (0.361), significantly surpassing \textit{Supercategory} and \textit{Random}, despite the lower scores among the EC metrics. We attribute this result to the high specificity leading to more simplistic communication strategies, characterized by a more limited and predictable set of symbols, making it easier to translate.

In addition, the exact match metrics (BLEU, METEOR, ROUGE-L) are strongly correlated with each other ($\rho = 0.72$), while they are slightly correlated with the CLIP score ($\rho = 0.26$), which provides semantic similarity between the candidate translation and the image itself. We attribute this to the inductive bias inherent in the experiment, as the ground truth are the images' captions written by annotators, and each image can be described in many ways. In spite of this, we observe a stronger correlation with \textit{Random} ($\rho = 0.37$) than with any of the other complexities. 

Finally, Figure \ref{fig:correlation_matrix_one_column}'s correlation matrix offers insights into how translation metrics interrelate across different setups and seeds. The matrix reveals that results for the same test set and game type are not correlated among all ECs, suggesting that the translation system captures different nuances from each EC. This lack of correlation indicates that each EC produces unique linguistic structures, highlighting the diversity in the EC developed by the agents.

\begin{figure}[h!]
    \centering
    \includegraphics[width=0.8\linewidth]{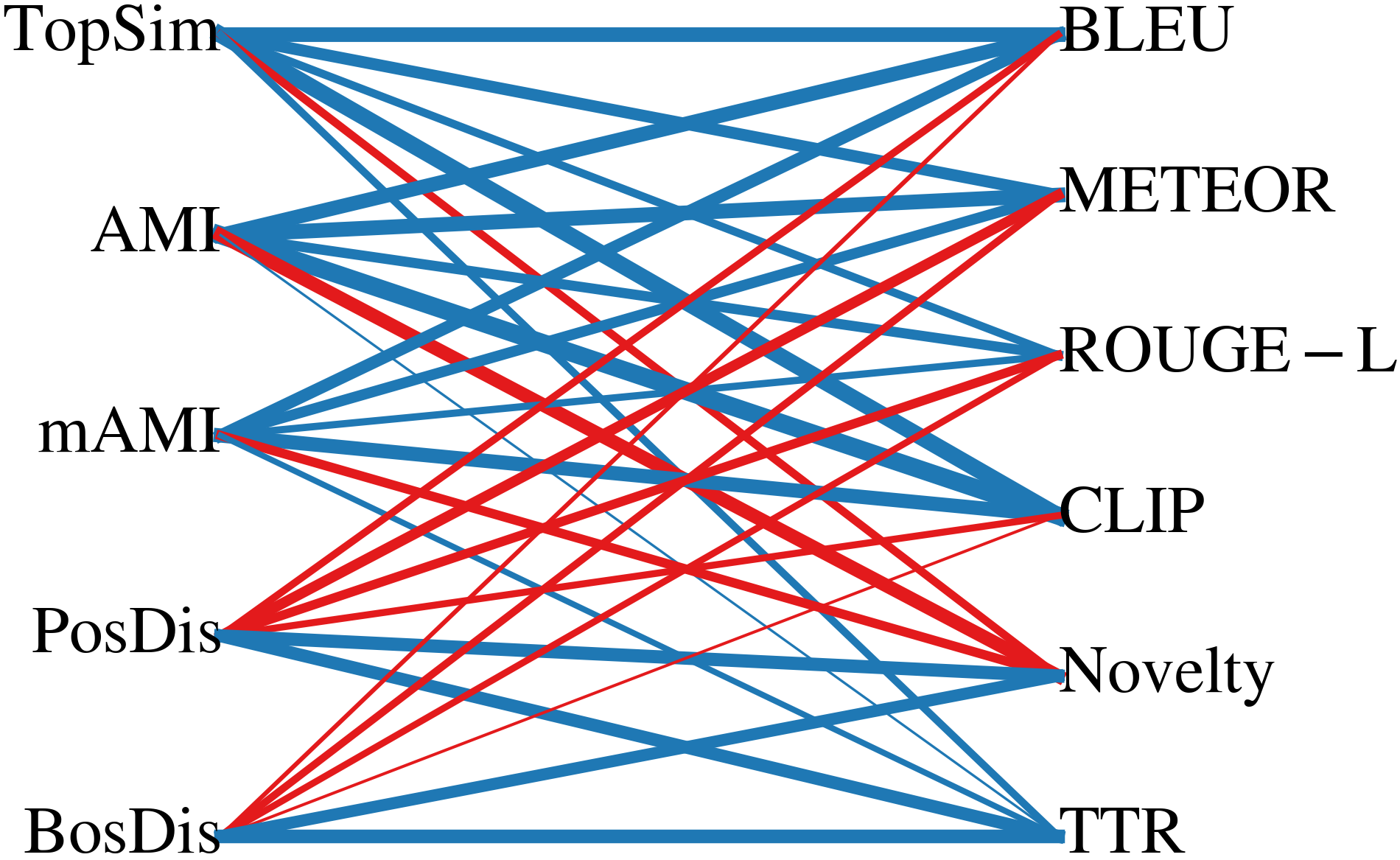}
  \caption{Bipartite network depicting the relationships between EC compositionality metrics and translatability metrics. Edge thickness is proportional to the absolute Pearson correlation (blue for positive, red for negative).}
\label{fig:CompUNMTcorr}
\end{figure}

\paragraph{Compositionality and Translatability} 
Figure~\ref{fig:CompUNMTcorr} illustrates the relationships between various EC compositionality measures and translation quality. Notably, \textbf{AMI} shows positive correlations with BLEU and METEOR but negative associations with Novelty, suggesting that messages with a close conceptual mapping tend to be easier to translate. Meanwhile, translation metrics compare generated translations to ground-truth captions; thus, a more ``novel'' expression naturally scores lower. In contrast, \textbf{BosDis} and \textbf{PosDis} focus on how strongly each symbol (or position) encodes a single attribute, offering a view of per-token ``disentanglement''. These metrics correlate negatively with standard translation scores but positively with Novelty. One possibility is that strictly mapping attributes to individual symbols results in messages with rich details and emphases, making translation more challenging, while simultaneously driving higher Novelty by introducing more unique expressions.
Additionally, \textbf{TopSim} is positively correlated with METEOR, suggesting that images with similar representations produce messages closer in structure, leading to greater translation predictability.
Overall, these findings suggest a connection between compositionality and translatability.

\section{Conclusion}

This research introduces a novel application of UNMT to translate AI agents' EC emerged in referential games into natural language. Our study demonstrates UNMT's potential to translate EC without parallel data and examines the impact of game complexity on EC translatability. Our findings indicate that different complexities produce distinct communication patterns. For instance, the \textit{Inter-Category} game generated EC that achieved significantly higher translation accuracy than the \textit{Supercategory} game, which often had lower scores despite its complexity. This suggests that not all complexities foster equally translatable communications.

Contrary to the initial hypothesis that more complex setups result in more detailed communications that are easier to translate, our empirical evidence shows otherwise. EC emerging from the \textit{Category} game, characterized by high contextual complexity, resulted in more pragmatic communication with significantly lower VU and Entropy while exhibiting high game accuracy. In contrast, the \textit{Supercategory} game, with slightly less contextual complexity, produced richer communication. We attribute this phenomenon to the agents' optimization for efficiency, where fine-grained distinctions encourage the use of low-level features. However, in terms of translatability, the \textit{Category} game achieved higher translation scores than the \textit{Supercategory} game, suggesting that a more pragmatic protocol facilitates translation.

Moreover, our analysis of compositionality and translation quality leads to a key hypothesis:  
while EC messages that closely reflect underlying concepts are more straightforward to translate,  
greater symbol-level detail increases novelty but also makes translation more challenging.

\subsection{Limitations and Future Work}
This study provides valuable insights into translating EC using UNMT. However, several limitations should be acknowledged. First, the number of distractors adds a layer of complexity that was not fully explored.
Second, sharing a pre-trained visual module may limit generality. Future work could explore separate or scratch-trained modules to assess their impact on EC translatability.
Third, the communication channel, vocabulary size, and messages length were fixed based on preliminary experiments. While these configurations were chosen to optimize performance, they may not capture the full range of possible communication strategies. Before translating, we analyzed many such configurations, ultimately selecting the setups with the best results.

Beyond limitations, the study suggests several future research avenues. EC is a rich field, characterized by a variety of techniques and strategies that form unique ECs. These range from multi-agent collaboration \cite{Michel2023RevisitingPI}, to bidirectional communication \cite{nikolaus2023emergent}, and reconstruction objectives \cite{chaabouni2021emergent}. Future research should enhance UNMT techniques to more effectively capture the subtleties of these ECs \cite{chauhan2022fully, amani2024symbolic} and assess faithfulness by having the Receiver evaluate how well the back-translation (EC→EN→EC) captures the original message's nuanced details. 
Finally, success in unsupervised translation of EC can further motivate similar attempts to translate other communication systems, such as animal communication \cite{goldwasser2024theory}.

\section*{Acknowledgements}

This research was supported by grant no.\ 2022330 from the United States - Israel Binational Science Foundation (BSF), Jerusalem, Israel.
IL, BC, and YB were supported by 
the Israel Science Foundation (grant no.\ 448/20), an Azrieli Foundation Early Career Faculty Fellowship, and an AI Alignment grant from Open Philanthropy. OP was funded by Project CETI via grants from Dalio Philanthropies and Ocean X; Sea Grape Foundation; Virgin Unite and Rosamund Zander/Hansjorg Wyss through The Audacious Project: a collaborative funding initiative housed at TED.
RM was supported by the Skillman chair.

\bibliography{Unsupervised_Translation_of_Emergent_Communication}

\appendix

\section{Images Dataset} \label{sec:appendix_dataset}

\begin{figure*}[ht]
    \centering
    \includegraphics[width=\linewidth]{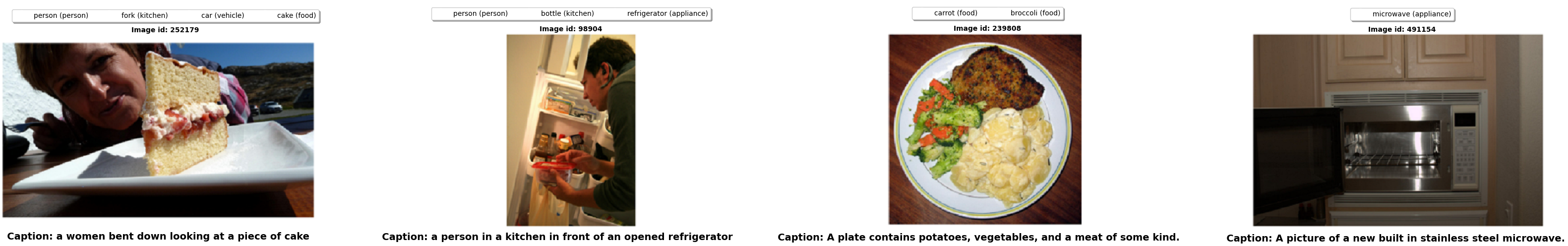}
    \caption{Images from the MSCOCO dataset along with their captions, categories, and supercategories. Annotated with NL, the images illustrate the complexity of the objects and scenes in the dataset.}
    \label{fig:dataset_samples}
\end{figure*}

For this study, we employed the MSCOCO (Microsoft Common Objects in Context) dataset \cite{Lin2014MicrosoftCC}, which contains 118,287 training images, 5,000 validation images, and 40,504 test images. Each image is annotated with five different captions, making it an excellent resource for tasks involving image captioning, translation, and language understanding. A wide range of scenarios and contexts are provided in the dataset, including 80 object categories and 12 supercategories, representing rigorous abstract NL image understanding, as shown in Figure~\ref{fig:dataset_samples}.

\subsection{Caption Statistics} \label{ssc:appendix_captions_statistics}

The captions in the MSCOCO dataset are rich in vocabulary and structure, providing a robust foundation for training and evaluating language models. Additionally, assist us in injecting prior knowledge of image descriptions into Unsupervised Machine Translation. Table \ref{tab:caption_stats} provides some statistics about the captions in the dataset.

\begin{table}[ht]
    \centering
    \begin{tabular}{lr}
        \toprule
        \textbf{Statistic} & \textbf{Value} \\
        \midrule
        Total number of captions & 591,835 \\
        Average caption length & 10.5 words \\
        Vocabulary size & 28,000 words \\
        Most frequent word & ``a'' \\
        \bottomrule
    \end{tabular}
    \caption{Statistics of captions in the MSCOCO dataset.}
    \label{tab:caption_stats}
\end{table}

\subsection{Access and Licensing}

The MSCOCO dataset is publicly available and can be accessed through the official website\footnote{\url{http://cocodataset.org/}}. The dataset is licensed for research purposes.

\section{Unsupervised Machine Translation} \label{ssc:Unsupervised_Machine_Translation}
Machine Translation (MT) has historically been constrained by its dependency on large-scale parallel corpora. However, the availability of such datasets is uneven across languages, with many lacking high-quality parallel texts, a limitation that is particularly acute in AI agents' EC. To address these challenges, Unsupervised Machine Translation (UMT) has emerged as a transformative approach, leveraging the abundance of available monolingual data.

UMT began by employing statistical decipherment methods, viewing the source language as an encoded script produced through a noisy channel. Early methods focused on assigning probabilities to potential translations in the target language \cite{ravi2013scalable, klementiev2012toward}. These initial efforts laid the groundwork for more sophisticated techniques, including both word-based and phrase-based statistical machine translation (SMT) methods \cite{koehn2003statistical, artetxe_etal_2018_unsupervised}. Despite this progress, deep learning has further revolutionized UMT by allowing for more complex and nuanced translation models \cite{yang_etal_2018_unsupervised, 
Artetxe2017}.
%hoshen_wolf_2018_non,
Adversarial training and iterative back-translation have proven particularly effective. In adversarial approaches \cite{Artetxe2017_mapping}, language embeddings are aligned across languages, effectively ``fooling'' a discriminator into not being able to distinguish the languages.

\textbf{Iterative back-translation} enhances translation accuracy by alternately improving translations in both directions between the source and target, by facilitating a continuous refinement cycle between two languages without parallel data. For example, an English sentence \( x \) is translated into Spanish yielding \( v(x) \), which is then back-translated into English \( u(v(x)) \). If the models are robust, this process should reconstruct \( x \), validating the translation's accuracy.

\textbf{Denoising auto-encoding} enhances model robustness by training to recover original texts from their noised versions, focusing on maintaining linguistic integrity despite disruptions. This process is pivotal in ensuring that UMT models remain effective even in linguistically noisy environments.

Hybrid models combining the strengths of unsupervised SMT and UNMT have also been explored, yielding further improvements in translation quality \cite{lample_etal_2018_phrase}. Despite these advances, challenges persist, particularly when source and target are highly dissimilar or when training across different domains, which can significantly degrade performance \cite{marchisio_etal_2020_unsupervised}.

Recently, \citet{goldwasser2024theory} motivated by the idea to communicate with animals, proposed a theoretical framework for analyzing UMT without parallel data.
According to this framework, successful UMT may be dependent on the complexity and commonality of the languages, suggesting that UMT could one day bridge gaps between human and non-human communication.

\begin{figure*}[ht]
    \centering
    \includegraphics[width=\linewidth]{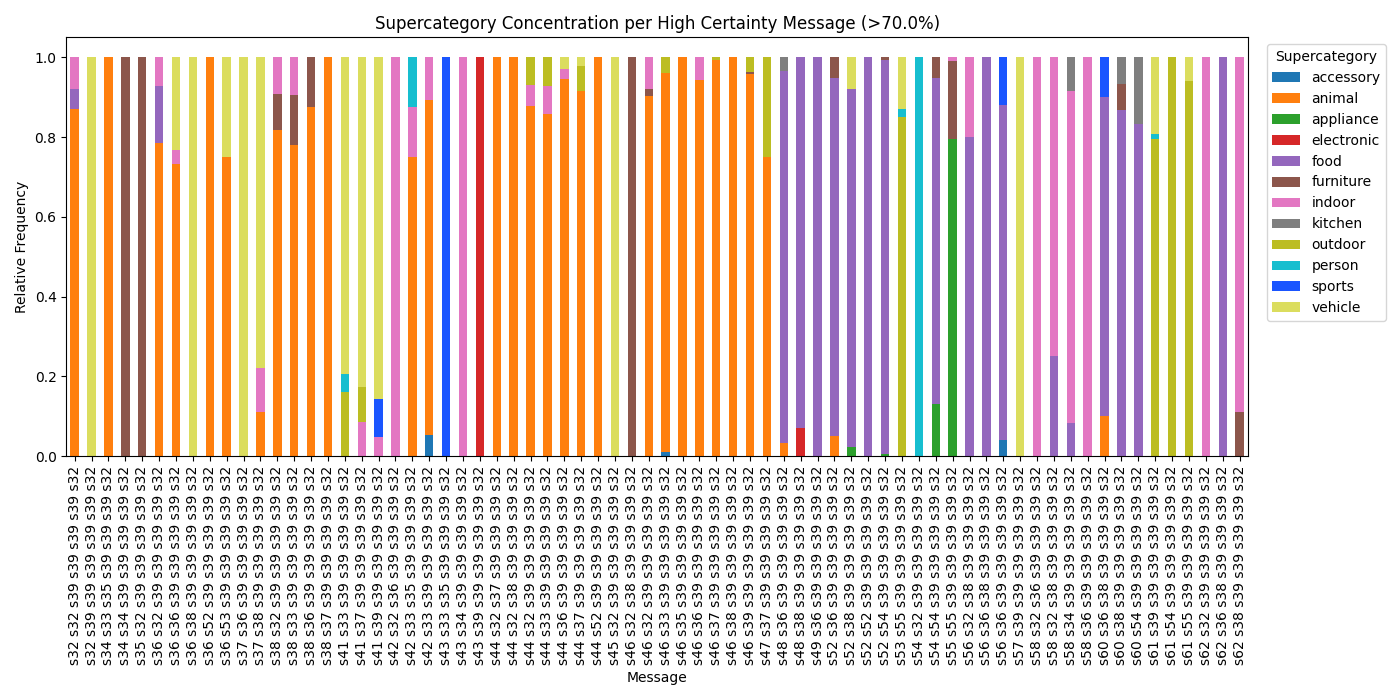}
    \caption{Distribution of image concepts across messages. The x-axis represents messages, and the bar colors represent the message’s assigned image concept distribution. Homogeneity in bars indicates that images with the same concept are being assigned by our agents to the same message.}
    \label{fig:concept_distribution}
\end{figure*}

\section{Hardware and Experiment Times}\label{sec:appendix_hardware_and_times}

The experiments were conducted on a system equipped with four NVIDIA RTX A4500 GPUs. Each GPU has 20470 MiB of memory and operates with a power usage of 200W.

The average training time for each model configuration and game type in the referential games was approximately 24 hours, whereas the training for each translation model lasted about 48 hours.

\section{Message to NL Concept Classification} \label{sec:appendix_ec_classification_2_NL_concepts}

The motivation for examining message compositionality in emergent communication (EC) is to understand how EC protocols can be mapped to NL concepts. As highlighted in \citet{carmeli2024concept}, EC compositionality can be evaluated by mapping EC words to NL concepts.
To explore the possibility of classifying EC messages into images' concepts, we used the following setup:
\begin{itemize}
    \item \textbf{Dataset}: MSCOCO, filtered to images with only one concept to facilitate classification.
    \item \textbf{Game}: Inter-category discrimination game (See Section \ref{ssec:complexities_definitions}).
    \item \textbf{Architecture}: LSTM + ResNet.
    \item \textbf{Communication Channel}:
        % \item Vocabulary size: 64 symbols
        % \item Message length: 7, including \textit{EOS} token
        % \item Quantization
         Vocabulary size of 64 symbols, with message length of 7, including \textit{EOS} token.
\end{itemize}

The Task is to classify EC messages to the input image concepts. It is followed by the motivation to understand how EC describes images and whether it is concept-wise.
We used a simple classifier where the most frequent concept in a message is considered as \( p(y|x) \).

The key observations are:
\begin{itemize}
    \item Messages serve as NL concept clusters.
    \item Images with the same concepts are assigned to the same message.
    \item \textbf{F-1 Score}: 0.643 among 12 categories.
\end{itemize}

This experiment provides insights into how EC describes images. The classifier is based on the most frequent concept in a message, where multiple images assigned to the same message contain one concept.

The plot in Figure \ref{fig:concept_distribution} shows the message's assigned image concept distribution, indicating homogeneity, which suggests that our EC protocol effectively describes concepts.

\begin{itemize}
    \item \textbf{Ratio of Messages where a Single Supercategory Dominates}:
    \begin{itemize}
        \item Ratio ($>$ 50\%): 0.67
        \item Ratio ($>$ 70\%): 0.45
        \item Ratio ($>$ 90\%): 0.23
    \end{itemize}
\end{itemize}

Although EC is a fascinating phenomenon, it remains an ongoing area of research. Some researchers have coined the term \textit{message collapse} to describe the situation where different images are mapped to the same message despite lacking observable similarity. This occurrence, where agents still succeed in the game, could explain why the distribution of concepts in our graph is not completely homogeneous.
The relatively high ratios of messages dominated by a single supercategory suggest that our EC captures image concepts, indicated by the observation that images with shared concepts tend to be mapped to the same messages. 

\section{Evaluation Metrics} \label{sec:appendix_evaluation_metrics}

\subsection{Metrics for Emergent Communication Games} \label{ssec:appendix_evaluation_metrics_EC}

\begin{enumerate}

\item  \textbf{Game Accuracy (ACC)}:
\[
\text{ACC\_$i$} = \frac{\text{Number of Correct Predictions}}{\text{Total Number of Predictions}}
\]
   Game accuracy measures how many correct identifications the agents make, and $i$ represents how many candidates they must choose from.

    \item \textbf{Vocabulary Usage (VU)}: \[
\text{VU} = \frac{|\{s_i \mid \text{count}(s_i) > 0\}|}{|S|}
\]
where \(|\{s_i \mid \text{count}(s_i) > 0\}|\) is the number of symbols used in messages and \(|S|\) is the total vocabulary size.

 \item \textbf{Message Entropy (ME)}:
\[
\text{ME} = -\sum_{m \in M} P(m) \log P(m)
\]
where \(M\) is the set of messages and \(P(m)\) is the probability of message \(m\).

\item \textbf{Message Novelty (MN)}:
\[
\text{MN} = \frac{|\{m \in M_{\text{test}} \mid m \notin M_{\text{train}}\}|}{|M_{\text{test}}|}
\]
where \(M_{\text{test}}\) is the set of messages in the test set and \(M_{\text{train}}\) is the set of messages in the training set.

\item \textbf{Topographic Similarity (TopSim)}:
\[
\text{TopSim} = \rho(d_{\text{input}}, d_{\text{message}})
\]
where \(\rho\) is the Spearman correlation between \(d_{\text{input}}\) and \(d_{\text{message}}\), with \(d_{\text{input}}\) representing the Euclidean distance in the input space and \(d_{\text{message}}\) representing the edit distance between corresponding messages. TopSim captures the alignment of distances in input and message spaces.

\item \textbf{Bag-of-Symbols Disentanglement (BosDis)}:
\item \textbf{Bag-of-Symbols Disentanglement (BosDis)}:
\[
  \text{BosDis} 
  = \sum_{j=1}^{|V|} 
  \frac{
    \Bigl(
    MI\bigl(s_j;\,a\bigr)
    \;-\;
    MI\bigl(s_j;\,b\bigr)
    \Bigr)
  }{
    H(s_j)
  }
\]

where \(s_j\) is the \(j\)-th symbol in the vocabulary \(V\), \(MI(\cdot;\cdot)\) denotes mutual information, and \(H(\cdot)\) denotes entropy. We compare a per-symbol measure of how much more strongly that symbol encodes one attribute \(a\) over the other \(b\).

\item \textbf{Positional Disentanglement (PosDis)}:
    PosDis follows the same logic as BosDis, but instead of looking at the MI of each symbol in the vocabulary, it focuses on each specific position in the message.

\item \textbf{Adjusted Mutual Information (AMI)}:
\[
\operatorname{AMI}(U, V) = \frac{\operatorname{MI}(U, V) - \mathbb{E}\left[\operatorname{MI}(U, V)\right]}{\max(H(U), H(V)) - \mathbb{E}\left[\operatorname{MI}(U, V)\right]}
\]
where \(I(U, V)\) is the mutual information between message clusters \(U\) and ground-truth labels \(V\), and \(\mathbb{E}[\operatorname{MI}(U, V)]\) is the expected MI under random assignment. AMI evaluates how well messages align with underlying image concepts, while adjusting for randomness.

\item \textbf{Multi-Concept AMI}:
    \[
    \operatorname{mAMI} = \frac{1}{|\mathcal{C}'|} \sum_{c \in \mathcal{C}'}\operatorname{AMI}_c(U, V)
    \]
    where \(c\) is concept and \(\mathcal{C}'\) is the set of valid concepts.

\end{enumerate}

\begin{figure*}[ht]
    \centering
    \includegraphics[width=\linewidth]
   {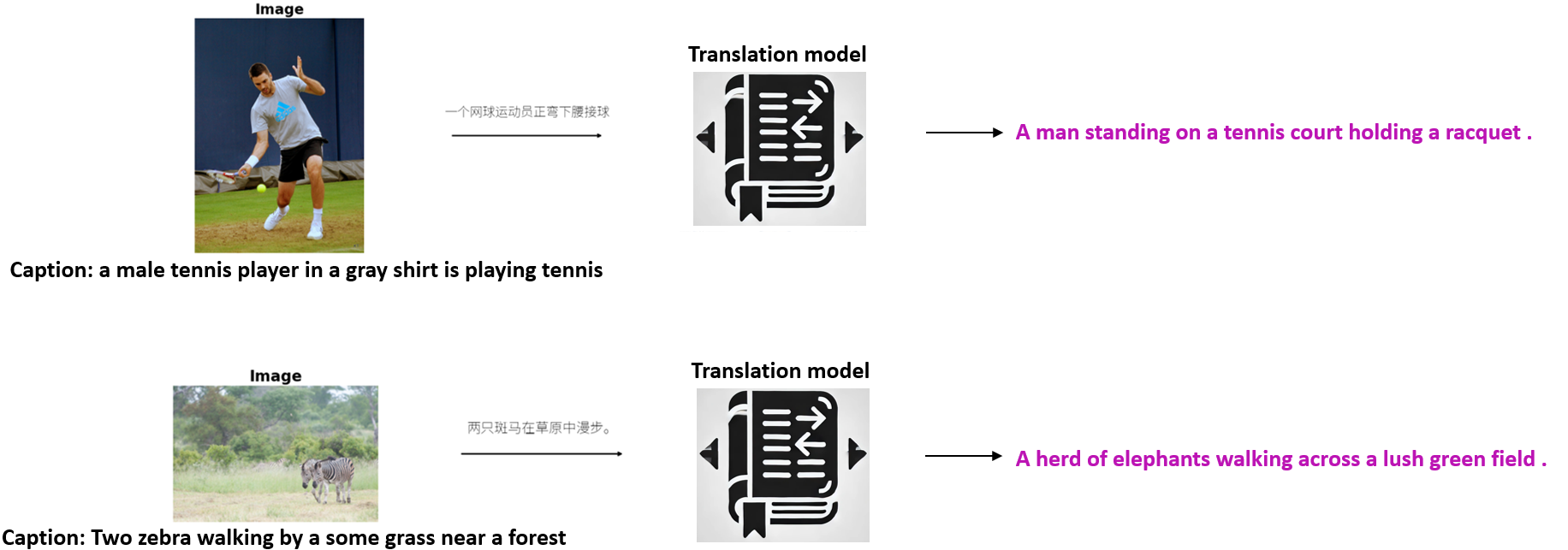}
    \caption{Examples of Chinese UNMT. Chinese captions and their translations are shown alongside the corresponding images.}
    \label{fig:chinese_unmt_examples}
\end{figure*}

\subsection{Metrics for UNMT} \label{ssec:appendix_evaluation_metrics_UNMT}

\begin{enumerate}

\item  \textbf{BERTScore}:
\[
\text{BERTScore}(R, C) = \frac{1}{|R|} \sum_{r \in R} \max_{c \in C} \text{cos}(e_r, e_c)
\]
where \(R\) is the set of reference embeddings, \(C\) is the set of candidate embeddings, and \(\text{cos}(e_r, e_c)\) is the cosine similarity between embeddings \(e_r\) and \(e_c\).

    \item \textbf{BLEU Score}:
\[
\text{BLEU} = \exp\left( \min\left(1 - \frac{|r|}{|c|}, 0\right) + \sum_{n=1}^{N} w_n \log p_n \right)
\]
where \(|r|\) is the length of the reference, \(|c|\) is the length of the candidate translation, \(w_n\) are the weights, and \(p_n\) is the precision of n-grams.

\item \textbf{Perplexity (PPL)}:
\[
\text{PPL} = \exp\left( -\frac{1}{N} \sum_{i=1}^{N} \log P(w_i \mid w_1, \ldots, w_{i-1}) \right)
\]
where \(N\) is the total number of words, and \(P(w_i \mid w_1, \ldots, w_{i-1})\) is the probability of word \(w_i\) given the previous words.

\item \textbf{METEOR}:
\[
\text{METEOR} = F_{\text{mean}} \cdot (1 - \text{Penalty})
\]
METEOR considers stemming and synonymy by using WordNet to account for matches beyond exact match, including stem and synonym matches. Each matched word is classified into one of these categories, and the total number of matches \(m\) is adjusted accordingly.

\item \textbf{CLIP Score}:
\[
\text{CLIP Score}(I, T) = \frac{1}{|T|} \sum_{t \in T} \text{cos}(e_t, e_i)
\]
where \(I\) is the set of image embeddings, \(T\) is the set of translation candidate embeddings, \(e_t\) is the embedding of the translation candidate, and \(e_i\) is the embedding of the corresponding image. All embeddings are obtained from the CLIP multi-modal model.

\item \textbf{Novelty Score}:
\[
\text{Novelty} = \frac{1}{|T|} \sum_{t \in T} I(t \notin D)
\]
where \(T\) is the set of n-grams in the translated texts, \(D\) is the set of n-grams in the training corpus, and \(I\) is an indicator function that is 1 if \(t\) is not in \(D\) and 0 otherwise. This metric evaluates the proportion of novel n-grams in the translations that do not appear in the training data.

    \item \textbf{Jaro Similarity}:
    \[
    \text{Jaro}(s_1, s_2) = \frac{1}{3} \left( \frac{m}{|s_1|} + \frac{m}{|s_2|} + \frac{m - t}{m} \right)
    \]
    where \(s_1\) and \(s_2\) are the input strings, \(m\) is the number of matching characters, and \(t\) is the transpositions number.

    \item \textbf{Text-Type Ratio (TTR)}:
    \[
    \text{TTR} = \frac{\text{Number of Unique Words}}{\text{Total Number of Words}}
    \]
    measures the lexical diversity within the translated texts.

\end{enumerate}

\section{Chinese UNMT} \label{sec:appendix_chinese_translation}

Before translating EC, we calibrated our expectations by evaluating how UNMT performs on a clear, interpretable and more complex language than EC. For this experiment, we collected a corpus of 120K English image captions \cite{Lin2014MicrosoftCC} and a 30K Chinese subset of it \cite{li2019coco} and trained the UNMT system with the same model and parameters used in other experiments.
Table \ref{tab:unmt_overall_performance}, in the Chinese column, presents the metrics for the Chinese-to-English UNMT. Figure \ref{fig:chinese_unmt_examples} provides examples of translated captions.
The Chinese UNMT results are summarized as follows:
\begin{itemize}
    % \item The translation system captured some context, though not verbatim.
    \item The UNMT system's ability to capture contextual meaning of source text, despite not always producing verbatim translations.
    \item Similar performance patterns between Chinese and EC translation strengthen our confidence in applying UNMT to emergent languages.
    % \item Translation metrics for Chinese were similar to those observed for EC.
    % \item Both unsupervised and semi-supervised approaches showed similar patterns to those in EC.
\end{itemize}

\section{Hyperparameters} \label{sec:appendix_hyperparameters}
The experiment can be divided into two main components, each with its own set of hyperparameters.
\par \textbf{EC Game} For each game type, referential games were trained with the following seeds: 31, 42, 123, 555, 999. Hyperparameters are reported in Section \ref{Implementation_details}. Regarding communication channel, the LSTM for both agents is with hidden sizes of 20 and embeddings of 50. 
\par \textbf{UNMT:} During Phase 2, the model was fine-tuned using a transformer architecture with 6 layers and 8 heads, embedding dimension of 1024, and Adam optimizer with a learning rate of 0.0001. The dropout rates for both standard and attention mechanisms were maintained at 0.1. For Phase 3, training objectives were autoencoding and backtranslation with additional word manipulation parameters—shuffle, dropout, and blanking, all set at 0.1, and translation sampling performed with a greedy decoder.

\begin{table*}[ht]
    \centering
    \begin{tabular}{lcccc}
        \toprule
        \textbf{Metric} & \textbf{Category} & \textbf{Supercategory} & \textbf{Random} & \textbf{Inter-category} \\
        \midrule
        BLEU & 39.31 {\scriptsize\(\pm\) 6.47} & 14.55 {\scriptsize\(\pm\) 11.91} & 27.78 {\scriptsize\(\pm\) 10.07} & 26.77 {\scriptsize\(\pm\) 15.79} \\
        BERTScore & 0.960 {\scriptsize\(\pm\) 0.0005} & 0.941 {\scriptsize\(\pm\) 0.0111} & 0.958 {\scriptsize\(\pm\) 0.0038} & 0.951 {\scriptsize\(\pm\) 0.0112} \\
        METEOR & 0.554 {\scriptsize\(\pm\) 0.049} & 0.278 {\scriptsize\(\pm\) 0.154} & 0.444 {\scriptsize\(\pm\) 0.108} & 0.410 {\scriptsize\(\pm\) 0.160} \\
        ROUGE-L & 0.588 {\scriptsize\(\pm\) 0.023} & 0.420 {\scriptsize\(\pm\) 0.094} & 0.505 {\scriptsize\(\pm\) 0.072} & 0.494 {\scriptsize\(\pm\) 0.088} \\
        \bottomrule
    \end{tabular}
    \caption{Performance Metrics for EN $\rightarrow$ EC Translation across Game Types. Mean \(\pm\) standard error are reported.}
    \label{tab:game_performance_metrics_EN2EC}
\end{table*}

\section{EN to EC Translation}\label{sec:appendix_EN2EC_analysis}
Given the iterative back-translation process at the core of our UNMT training, the model inherently learns to translate in both directions—EC \(\rightarrow\) EN and EN \(\rightarrow\) EC.
The EN \(\rightarrow\) EC translation performance presented in Table~\ref{tab:game_performance_metrics_EN2EC} reinforces our core assertion that emergent protocols optimized for pragmatic clarity are inherently more translatable. In particular, the Category game yields significantly higher BLEU (39.31) compared to other complexities, indicating a strong alignment between the emergent messages and their NL counterparts. Conversely, the Supercategory setting, characterized by richer but more variable communication, shows substantially lower scores (e.g., BLEU: 14.55; METEOR: 0.278), suggesting that increased structural diversity—while expressive—compromises direct translatability. The intermediate performance observed in the Random and Inter-category games further underscores the delicate balance between expressiveness and translatability. We excluded the EOS token from our calculations, to ensure that these metrics precisely capture translation quality without any bias.

\section{Examples} \label{sec:appendix_king_solomon_translations}
The tables presented below offer an overview of illustrative translation examples selected from the top 1000 BLEU scores in our test translations. Each table corresponds to a specific game type—Random, Category, Supercategory, and Inter-category. This selection is curated to showcase the breadth and depth of linguistic transformations observed from translations, providing insights into how EC vocabulary adapts to varying contextual demands.

%%%%%%%%% Random %%%%%%%%%%
\begin{table*}
\centering
\textbf{Game Type: Random}
\small
\begin{tabular}{p{8.5cm}p{8.5cm}}
\toprule
\textbf{Image + Caption} & \textbf{EC message + Translation} \\
\midrule
\includegraphics[width=4cm]{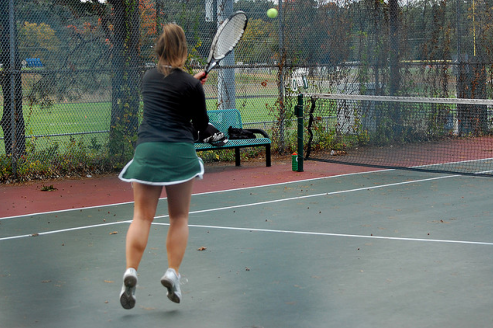}
\newline
A woman on a court with a tennis racket. & 
s35 s40 s40 s40 s40 s40 s32 \\
& \textcolor{purple}{\textit{A man on a court with a tennis racket.}} \\
\midrule
\includegraphics[width=4cm]{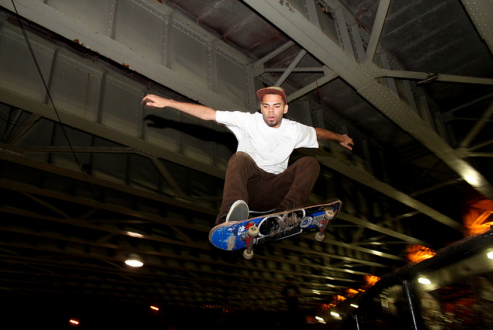}
\newline
A young man is jumping in the air on a skateboard. & 
s3 s43 s43 s42 s42 s42 s32 \\
& \textcolor{purple}{\textit{A skateboarder is jumping in the air on a ramp.}} \\
\midrule
\includegraphics[width=4cm]{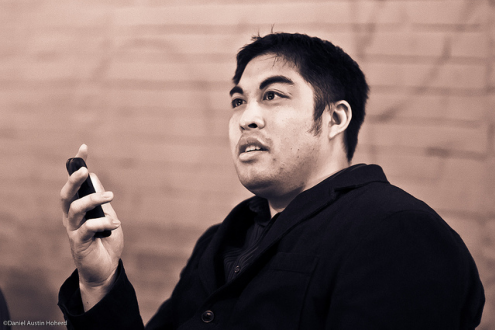}
\newline
A man in a black shirt is holding a cell phone. & 
s3 s58 s43 s42 s42 s42 s32 \\
& \textcolor{purple}{\textit{A person in a blue shirt is holding a cell phone.}} \\
\midrule
\includegraphics[width=4cm]{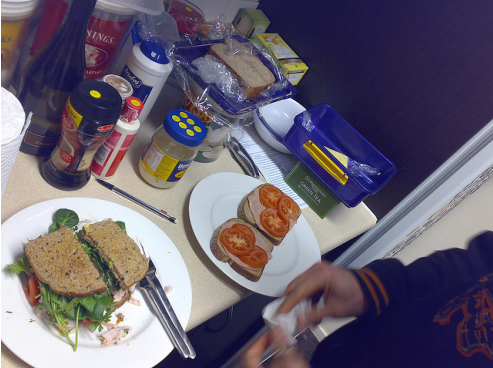}
\newline
A counter topped with a sandwich on a plate. & 
s16 s50 s42 s42 s42 s40 s32 \\
& \textcolor{purple}{\textit{A white plate topped with a sandwich on a table.}} \\
\midrule
\includegraphics[width=4cm]{Figures/Translations_Examples/unsupervised_227891.png}
\newline
A white table with a laptop and another monitor. & 
s13 s60 s44 s40 s40 s40 s32 \\
& \textcolor{purple}{\textit{A table with a laptop and a mouse.}} \\
\bottomrule
\end{tabular}
\caption{Examples of emergent communication messages and their translations for the Random game type, annotated with images and captions.}
\label{tab:random_translations}
\end{table*}

%%%%%%%%% Category %%%%%%%%%%
\begin{table*}
\centering
\textbf{Game Type: Category}
\small
\begin{tabular}{p{8.5cm}p{8.5cm}}
\toprule
\textbf{Image + Caption} & \textbf{EC message + Translation} \\
\midrule
\includegraphics[width=4cm]{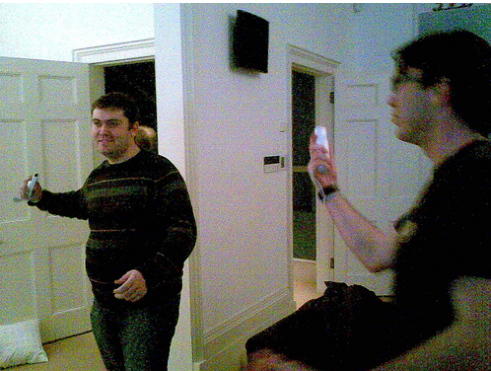}
\newline
Two men are playing a video game together. &
s44 s53 s53 s53 s53 s53 s32 \\
& \textcolor{purple}{\textit{Two men are playing a video game together.}} \\
\midrule
\includegraphics[width=4cm]{Figures/Translations_Examples/category_324349.png}
\newline
A man riding a skateboard through a pavilion. &
s53 s53 s53 s53 s53 s53 s32 \\ 
& \textcolor{purple}{\textit{A man riding a skateboard down a street.}} \\
\midrule
\includegraphics[width=4cm]{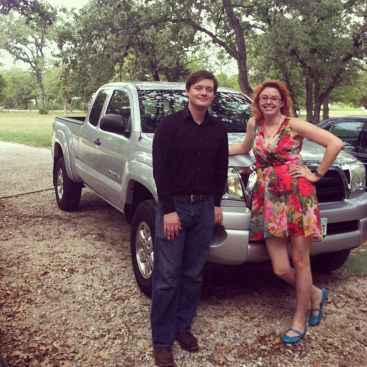}
\newline
Two people standing in front of a truck. &
s44 s52 s51 s53 s53 s53 s32 \\
& \textcolor{purple}{\textit{Two people are standing in front of a bus.}} \\
\midrule
\includegraphics[width=4cm]{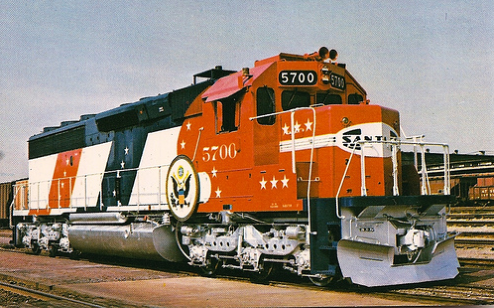}
\newline
A train is traveling down the railroad tracks. &
s32 s49 s49 s53 s53 s53 s32 \\
& \textcolor{purple}{\textit{A train is traveling down the tracks near a platform.}} \\
\midrule
\includegraphics[width=4cm]{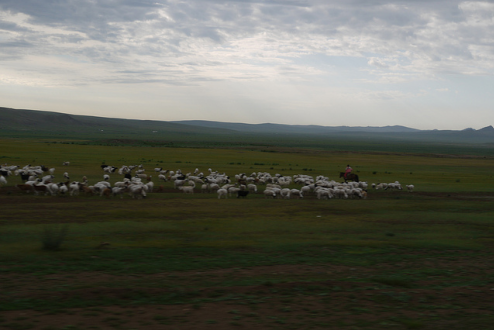}
\newline
A herd of sheep grazing in an open field &
s55 s53 s53 s53 s53 s53 s32 \\
& \textcolor{purple}{\textit{A herd of sheep grazing on a lush green field.}} \\
\bottomrule
\end{tabular}
\caption{Examples of emergent communication messages and their translations for the Category game type, annotated with images and captions.}
\label{tab:category_translations}
\end{table*}

%%%%%%%%% Supercategory %%%%%%%%%%
\begin{table*}
\centering
\textbf{Game Type: Supercategory}
\small
\begin{tabular}{p{8.5cm}p{8.5cm}}
\toprule
\textbf{Image + Caption} & \textbf{EC message + Translation} \\
\midrule
\includegraphics[width=4cm, height=5cm]{Figures/Translations_Examples/supercategory_151832.png}
\newline
A small bathroom with a toilet and a sink. &
s23 s7 s7 s7 s7 s7 s32 \\
& \textcolor{purple}{\textit{A bathroom with a toilet and a sink.}} \\
\midrule
\includegraphics[width=4cm, height=5cm]{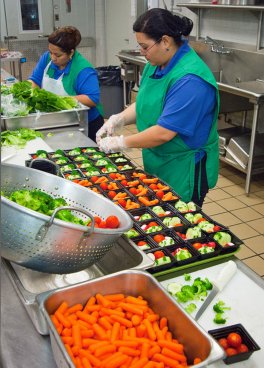}
\newline
A woman standing in a kitchen preparing food. &
s22 s30 s28 s20 s20 s20 s32 \\
& \textcolor{purple}{\textit{Man standing in a kitchen preparing food in a kitchen.}} \\
\midrule
\includegraphics[width=4cm]{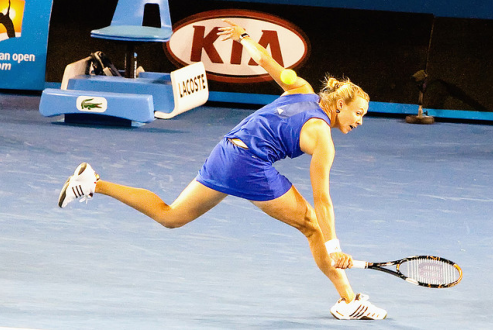}
\newline
A woman holding a tennis racquet on a tennis court. &
s10 s10 s8 s8 s40 s40 s32 \\
& \textcolor{purple}{\textit{Girl holding a tennis racket on a tennis court.}} \\
\midrule
\includegraphics[width=4cm]{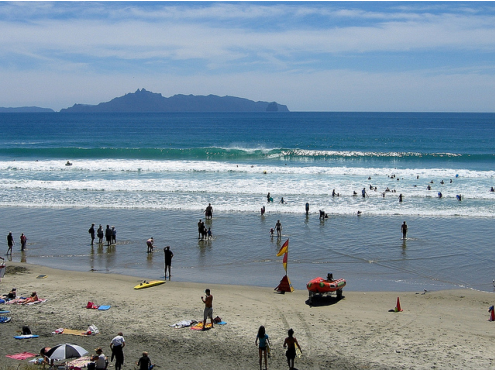}
\newline
A group of people walk on a beach. &
s23 s23 s23 s22 s6 s38 s32 \\
& \textcolor{purple}{\textit{A group of people and a dog on a beach.}} \\
\bottomrule
\end{tabular}
\caption{Examples of emergent communication messages and their translations for the Supercategory game type, annotated with images and captions.}
\label{tab:supercategory_translations}
\end{table*}

%%%%%%%%% Inter-category %%%%%%%%%%
\begin{table*}
\centering
\textbf{Game Type: Inter-category}
\small
\begin{tabular}{p{8.5cm}p{8.5cm}}
\toprule
\textbf{Image + Caption} & \textbf{EC message + Translation} \\
\midrule
\includegraphics[width=4cm]{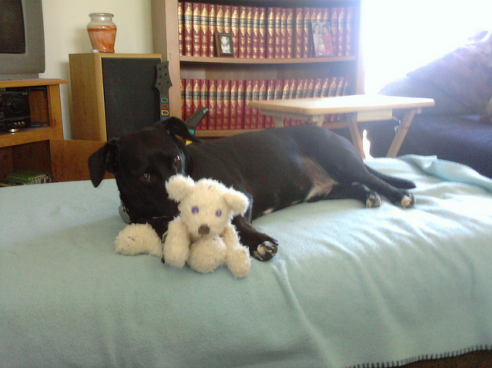}
\newline
A dog laying on a bed with a stuffed animal. &
s17 s19 s23 s23 s7 s7 s32 \\
& \textcolor{purple}{\textit{A child sitting on a bed with a stuffed animal.}} \\
\midrule
\includegraphics[width=4cm]{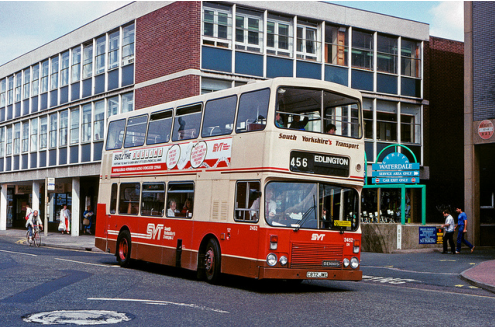}
\newline
A red and white double decker bus driving down a street. &
s56 s25 s25 s9 s9 s9 s32 \\
& \textcolor{purple}{\textit{A blue and white bus driving down a street.}} \\
\midrule
\includegraphics[width=4cm]{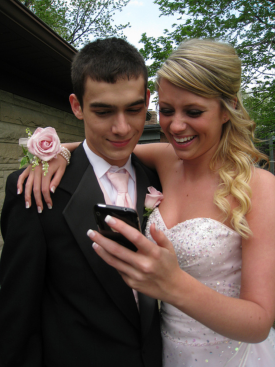}
\newline
A close up of two people looking at a cell phone. &
s11 s15 s15 s7 s7 s7 s32 \\
& \textcolor{purple}{\textit{A close up of a person holding a cell phone.}} \\
\midrule
\includegraphics[width=4cm]{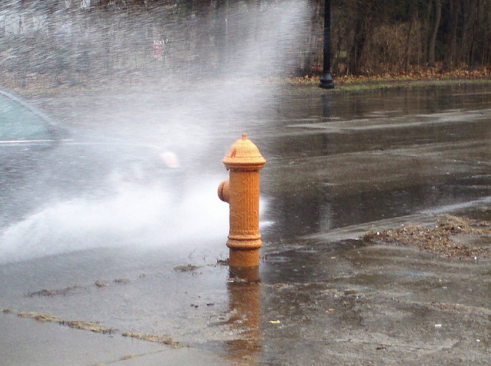}
\newline
A yellow fire hydrant that is spraying water. &
s42 s24 s8 s8 s8 s8 s32 \\
& \textcolor{purple}{\textit{A yellow fire hydrant is on a sidewalk.}} \\
\bottomrule
\end{tabular}
\caption{Examples of emergent communication messages and their translations for the Inter-category game type, annotated with images and captions.}
\label{tab:intercategory_translations}
\end{table*}

\end{document}